# A Bayesian Approach to Tackling Hard Computational Problems

**Eric Horvitz**
Microsoft Research
Redmond, WA 98052
horvitz@microsoft.com

**Yongshao Ruan**
Comp. Sci. & Engr.
Univ. of Washington
Seattle, WA 98195
ruan@cs.washington.edu

**Carla Gomes**
Dept. of Comp. Sci.
Cornell Univ.
Ithaca, NY 14853
gomes@cs.cornell.edu

**Henry Kautz**
Comp. Sci. & Engr.
Univ. of Washington
Seattle, WA 98195
kautz@cs.washington.edu

**Bart Selman**
Dept. of Comp. Sci.
Cornell Univ.
Ithaca, NY 14853
selman@cs.cornell.edu

**Max Chickering**
Microsoft Research
Redmond, WA 98052
dmax@microsoft.com

## Abstract

We describe research and results centering on the construction and use of Bayesian models that can predict the run time of problem solvers. Our efforts are motivated by observations of high variance in the time required to solve instances for several challenging problems. The methods have application to the decision-theoretic control of hard search and reasoning algorithms. We illustrate the approach with a focus on the task of predicting run time for general and domain-specific solvers on a hard class of structured constraint satisfaction problems. We review the use of learned models to predict the ultimate length of a trial, based on observing the behavior of the search algorithm during an early phase of a problem session. Finally, we discuss how we can employ the models to inform dynamic run-time decisions.

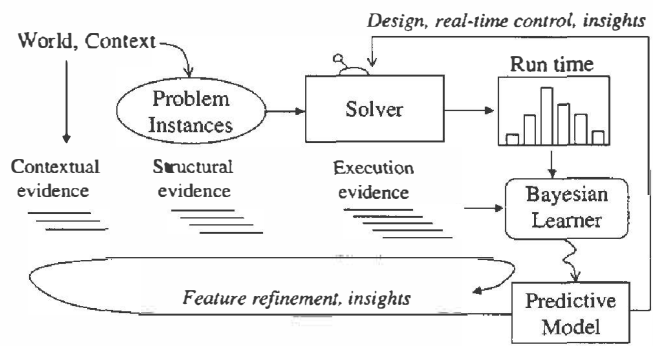

Figure 1: Bayesian approach to problem solver design and optimization. We seek to learn predictive models to refine and control computational procedures as well as to gain insights about problem structure and hardness.

## 1 Introduction

The design of procedures for solving difficult problems relies on a combination of insight, observation, and iterative refinements that take into consideration the behavior of algorithms on problem instances. Complex, impenetrable relationships often arise in the process of problem solving, and such complexity leads to uncertainty about the basis for observed efficiencies and inefficiencies associated with specific problem instances. We believe that recent advances in Bayesian methods for learning predictive models from data offer valuable tools for designing, controlling, and understanding automated reasoning methods.

We focus on using machine learning to characterize variation in the run time of instances observed in inherently exponential search and reasoning problems. Predictive models for run time in this domain could provide the basis for more optimal decision making at the microstructure of algorithmic activity as well as inform higher-level policies that guide the allocation of resources.

Our overall methodology is highlighted in Fig. 1. We seek to develop models for predicting execution time by considering dependencies between execution time and one or more classes of observations. Such classes include evidence about the nature of the generator that has provided instances, about the structural properties of instances noted before problem solving, and about the run-time behaviors of solvers as they struggle to solve the instances.

The research is fundamentally iterative in nature. We exploit learning methods to identify and continue to refine observational variables and models, balancing the predictive power of multiple observations with the cost of the real-time evaluation of such evidential dis-tinctions. We seek ultimately to harness the learned models to optimize the performance of automated reasoning procedures. Beyond this direct goal, the overall exploratory process promises to be useful for providing new insights about problem hardness.

We first provide background on the problem solving domains we have been focusing on. Then, we describe our efforts to instrument problem solvers and to learn predictive models for run time. We describe the formulation of variables we used in data collection and model construction and review the accuracy of the inferred models. Finally, we discuss opportunities for exploiting the models. We focus on the sample application of generating context-sensitive restart policies in randomized search algorithms.

## 2 Hard Search Problems

We have focused on applying learning methods to characterize run times observed in backtracking search procedures for solving NP-complete problems encoded as constraint satisfaction (CSP) and Boolean satisfiability (SAT). For these problems, it has proven extremely difficult to predict the particular sensitivities of run time to changes in instances, initialization settings, and solution policies. Numerous studies have demonstrated that the probability distribution over run times exhibit so-called *heavy-tails* [10]. Restart strategies have been used in an attempt to find settings for an instance that allow it to be solved rapidly, by avoiding costly journeys into a long tail of run time. Restarts are introduced by way of a parameter that terminates the run and restarts the search from the root with a new random seed after some specified amount of time passes, measured in choices or backtracks.

Progress on the design and study of algorithms for SAT and CSP has been aided by the recent development of new methods for generating hard random problem instances. Pure random instances, such as $k$-Sat, have played a key role in the development of algorithms for propositional deduction and satisfiability testing. However, they lack the structure that characterizes real world domains. Gomes and Selman [9] introduced a new benchmark domain based on *Quasigroups*, the Quasigroup Completion Problem (QCP). QCP captures the structure that occurs in a variety of real world problems such as timetabling, routing, and statistical experimental design.

A quasigroup is a discrete structure whose multiplication table corresponds to a Latin Square. A *Latin Square of order $n$* is an $n \times n$ array in which $n$ distinct symbols are arranged so that each symbol occurs once in each row and column. A partial quasigroup (or Latin Square) of order $n$ is an $n \times n$ array based on $n$ distinct symbols in which some cells may be empty but no row or column contains the same element twice. The Quasigroup Completion Problem (QCP) can be stated as follows: Given a partial quasigroup of order $n$ can it be completed to a quasigroup of the same order?

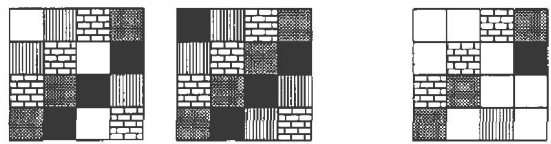

Figure 2: Graphical representation of the quasigroup problem. Left: A quasigroup instance with its completion. Right: A balanced instance with two holes per row/column.

QCP is an NP-complete problem [5] and random instances have been found to exhibit a peak in problem hardness as a function of the ratio of the number of uncolored cells to the total number of cells. The peak occurs over a particular range of values of this parameter, referred to as a region of *phase transition* [9, 2]. A variant of the QCP problem, *Quasigroup with Holes* (QWH) [2], includes only satisfiable instances. The QWH instance-generation procedure essentially inverts the completion task: it *begins* with a randomly-generated completed Latin square, and then erases colors or "pokes holes." Completing QWH is NP-Hard [2]. A structural property that affects hardness of instances significantly is the pattern of the holes in row and columns. Balancing the number holes in each row and column of instances has been found to significantly increase the hardness of the problems [1].

## 3 Experiments with Problem Solvers

We performed a number of experiments with Bayesian learning methods to elucidate previously hidden distinctions and relationships in SAT and CSP reasoners. We experimented with both a randomized SAT algorithm running on Boolean encodings of the QWH and a randomized CSP solver for QWH. The SAT algorithm was Satz-Rand [11], a randomized version of the Satz system of Li and Anbulagan [20]. Satz is the fastest known complete SAT algorithm for hard random 3-SAT problems, and is well suited to many interesting classes of structured satisfiability problems, including SAT encodings of quasigroup completion problems [10] and planning problems [17]. The solver is a version of the classic Davis-Putnam (DPLL) algorithm [7] augmented with one-step lookahead and a sophisti-cated variable choice heuristic. The lookahead operation is invoked at most choice points and finds any variable/value choices that would immediately lead to a contradiction after unit propagation; for these, the opposite variable assignment can be immediately made. The variable choice heuristic is based on picking a variable that if set would cause the greatest number of ternary clauses to be reduced to binary clauses. The variable choice set was enlarged by a noise parameter of 30%, and value selection was performed deterministically by always branching on 'true' first.

The second backtrack search algorithm we studied is a randomized version of a specialized CSP solver for quasigroup completion problems, written using the ILOG solver constraint programming library. The backtrack search algorithm uses as a variable choice heuristic a variant of the Brelaz heuristic. Furthermore, it uses a sophisticated propagation method to enforce the constraints that assert that all the colors in a row/column must be different. We refer to such a constraint as *alldiff*. The propagation of the *alldiff* constraint corresponds to solving a matching problem on a bipartite graph using a network-flow algorithm [9, 26, 24].

We learned predictive models for run-time, motivated by two different classes of target problems. For the first class of problem, we assume that a solver is challenged by an instance and must solve that specific problem as quickly as possible. We term this the *Single Instance* problem. In a second class of problem, we draw cases from a distribution of instances and are required to solve any instance as soon as possible, or as many instances as possible for any amount of time allocated. We call these challenges *Multiple Instance* problems, and the subproblems as the *Any Instance* and *Max Instances* problems, respectively.

We collected evidence and built models for CSP and Satz solvers applied to the QWH problem for both the Single Instance and Multiple Instances challenge. We shall refer to the four problem-solving experiments as CSP-QWH-Single, CSP-QWH-Multi, Satz-QWH-Single, and Satz-QWH-Multi. Building predictive Bayesian models for the CSP-QWH-Single and Satz-QWH-Single problems centered on gathering data on the probabilistic relationships between observational variables and run time for single instances with randomized restarts. Experiments for the CSP-QWH-Multi and Satz-QWH-Multi problems centered on performing single runs on multiple instances drawn from the same instance generator.

### 3.1 Formulating Evidential Variables

We worked to define variables that we believed could provide information on *problem-solving progress* for a period of observation in an early phase of runs that we refer to as the *observation horizon*. The definition of variables was initially guided by intuition. However, results from our early experiments helped us to refine sets of variables and to propose additional candidates.

We initially explored a large number of variables, including those that were difficult to compute. Although we planned ultimately to avoid the use of costly observations in real-time forecasting settings, we were interested in probing the predictive power and interdependencies among features regardless of cost. Understanding such informational dependencies promised to be useful in understanding the potential losses in predictive power with the removal of costly features, or substitution of expensive evidence with less expensive, approximate observations. We eventually limited the features explored to those that could be computed with low (constant) overhead.

We sought to collect information about base values as well as several variants and combinations of these values. For example, we formulated features that could capture higher-level patterns and dynamics of the state of a problem solver that could serve as useful probes of solution progress. Beyond exploring base observations about the program state at particular points in a case, we defined new families of observations such as first and second derivatives of the base variables, and summaries of the status of variables over time.

Rather than include a separate variable in the model for each feature at each choice point—which would have led to an explosion in the number of variables and severely limited generalization—features and their dynamics were represented by variables for their summary statistics over the observation horizon. The summary statistics included initial, final, average, minimum, and maximum values of the features during the observation period. For example, at each choice point, the SAT solver recorded the current number of binary clauses. The training data would thus included a variable for the average first derivative of the number of binary clauses during the observation period. Finally, for several of the features, we also computed a summary statistic that measured the number of times the sign of the feature changed from negative to positive or vice-versa.

We developed distinct sets of observational variables for the CSP and Satz solvers. The features for the CSP solver included some that were generic to any constraint satisfaction problem, such as the number of backtracks, the depth of the search tree, and the average domain size of the unbound CSP variables. Other features, such as the variance in the distribution of unbound CSP variables between different columns of the square, were specific to Latin squares. As we will see below, the inclusion of such domain-specific features was important in learning strongly predictive models. The CSP solver recorded 18 basic features at each choice point which were summarized by a total of 135 variables. The variables that turned out to be most informative for prediction are described in Sec. 4.1 below.

The features recorded by Satz-Rand were largely generic to SAT. We included a feature for the number of Boolean variables that had been set positively; this feature is problem specific in the sense that under the SAT encoding we used, only a *positive* Boolean variable corresponds to a *bound* CSP variable (*i.e.* a colored squared). Some features measured the current problem size (*e.g.* the number of unbound variables), others the size of the search tree, and still others the effectiveness of unit propagation and lookahead.

We also calculated two other features of special note. One was the logarithm of the total number of possible truth assignments (models) that had been ruled out at any point in the search; this quantity can be efficiently calculated by examining the stack of assumed and proven Boolean variable managed by the DPLL algorithm. The other is a quantity from the theory of random graphs called $\lambda$, that measures the degree of interaction between the binary clauses of the formula [23]. In all Satz recorded 25 basic features that were summarized in 127 variables.

### 3.2 Collecting Run-Time Data

For all experiments, observational variables were collected over an observational horizon of 1000 solver choice points. Choice points are states in search procedures where the algorithm assigns a value to variables heuristically, per the policies implemented in the problem solver. Such points do not include the cases where variable assignment is forced via propagation of previous set values, as occurs with unit propagation, backtracking, lookahead, and forward-checking.

For the studies described, we represented run time as a binary variable with discrete states short versus long. We defined short runs as cases completed before the median of the run times for all cases in each data set. Instances with run times shorter than the observation horizon were not considered in the analyses.

## 4 Models and Results

We employed Bayesian structure learning to infer predictive models from data and to identify key variables from the larger set of observations we collected. Over the last decade, there has been steady progress on methods for inferring Bayesian networks from data [6, 27, 12, 13]. Given a dataset, the methods typically perform heuristic search over a space of dependency models and employ a Bayesian score to identify models with the greatest ability to predict the data. The Bayesian score estimates $p(model|data)$ by approximating $p(data|model)p(model)$. Chickering, Heckerman and Meek [4] show how to evaluate the Bayesian score for models in which the conditional distributions are decision trees. This Bayesian score requires a prior distribution over both the parameters and the structure of the model. In our experiments, we used a uniform parameter prior. Chickering et al. suggest using a structure prior of the form: $p(model) = \kappa^{fp}$, where $0 < \kappa \leq 1$ and $fp$ is the number of free parameters in the model. Intuitively, smaller values of $\kappa$ make large trees unlikely a priori, and thus $\kappa$ can be used to help avoid overfitting. We used this prior, and tuned $\kappa$ as described below.

We employed the methods of Chickering et al. to infer models and to build decision trees for run time from the data collected in experiments with CSP and Satz problem solvers applied to QWH problem instances. We shall describe sample results from the data collection and four learning experiments, focusing on the CSP-QWH-Single case in detail.

### 4.1 CSP-QWH-Single Problem

For a sample CSP-QWH-Single problem, we built a training set by selecting nonbalanced QWH problem instance of order 34 with 380 unassigned variables. We solved this instance 4000 times for the training set and 1000 times for the test data set, initiating each run with a random seed. We collected run time data and the states of multiple variables for each case over an observational horizon of 1000 choice points. We also created a marginal model, capturing the overall run-time statistics for the training set.

We optimized the $\kappa$ parameter used in the structure prior of the Bayesian score by splitting the training set 70/30 into training and holdout data sets, respectively. We selected a kappa value by identifying a soft peak in the Bayesian score. This value was used to build a dependency model and decision tree for run time from the full training set. We then tested the abilities of the marginal model and the learned decision tree to predict the outcomes in the test data set. We computed a classification accuracy for the learned and marginal models to characterize the power of these models. The classification accuracy is the likelihood that the classifier will correctly identify the run time of cases in the test set. We also computed an average log score for the models.

Fig. 3 displays the learned Bayesian network for this dataset. The figure highlights key dependencies and variables discovered for the data set. Fig. 4 shows the decision tree for run time.

The classification accuracy for the learned model is 0.963 in contrast with a classification accuracy of 0.489 for the marginal model. The average log score of the learned model is -0.134 and the average log score of the marginal model was -0.693.

Because this was both the strongest and most compact model we learned, we will discuss the features it involves in more detail. Following Fig. 4 from left to right, these are:

*VarRowColumn* measures the variance in the number of uncolored cells in the QWH instance across rows and across columns. A low variance indicates the open cells are evenly balanced throughout the square. As noted earlier, balanced instances are harder to solve than unbalanced ones [1]. A rather complex summary statistic of this quantity appears at the root of the decision tree, namely the minimum of the first derivative of this quantity during the observation period. In future work we will be examining this feature carefully in order to determine why this particular statistic was most relevant.

*AvgColumn* measures the ratio of the number of uncolored cells and the number of columns or rows. A low value for this feature indicates that the quasigroup is nearly complete. The decision tree shows that a run is likely to be fast if the minimum value of this quantity over the entire observation period is small.

*MinDepth* is the minimum depth of all leaves of the search tree, and the summary statistic is simply the final value of this quantity. The third and fourth nodes of the decision tree show that short runs are associated with high minimum depth and long runs with low minimum depth. This may be interpreted as indicating the search trees for the shorter runs have a more regular shape.

*AvgDepth* is the average depth of a node in the search tree. The model discovers that short runs are associated with a high frequency in the change of the sign of the first derivative of the average depth. In other words, frequent fluctuations up and down in the average depth indicate a short run. We do not yet have an intuitive explanation for this phenomena.

*VarRowColumn* appears again as the last node in the decision tree. Here we see that if the maximum variance of the number of uncolored cells in the QWH instance across rows and columns is low (*i.e.*, the problem remains balanced) then the run is long, as might be expected.

### 4.2 CSP-QWH-Multi Problem

For a CSP-QWH-Multi problem, we built training and test sets by selecting instances of nonbalanced QWH problems of order 34 with 380 unassigned variables. We collected data on 4000 instances for the training set and 1000 instances for the test set.

As we were running instances of potentially different fundamental hardnesses, we normalized the feature measurements by the size of the instance (measured in CSP variables) *after* the instances were initially simplified by forward-checking. That is, although all the instances originally had the same number of uncolored cells, polynomial time preprocessing fills in some of the cells, thus revealing the true size of the instance.

We collected run time data for each instance over an observational horizon of 1000 choice points. The learned model was found to have a classification accuracy of 0.715 in comparison to the marginal model accuracy of 0.539. The average log score for the learned model was found to be -0.562 and the average log score for the marginal model was -0.690.

### 4.3 Satz-QWH-Single Problem

We performed analogous studies with the Satz solver. In a study of the Satz-QWH-Single problem, we studied a single QWH instance (bqwh-34-410-16). We found that the learned model had a classification accuracy of 0.603, in comparison to a classification accuracy of 0.517 for the marginal model. The average log score of the learned model was found to be -0.651 and the log score of the marginal model was -0.693.

The predictive power of the SAT model was less than that of the corresponding CSP model. This is reasonable since the CSP model had access to features that more precisely captured special features of quasigroup problems (such as balance). The decision tree was still relatively small, containing 12 nodes that referred to 10 different summary variables.

Observations that turned out to be most relevant for the SAT model included:

- The maximum number of variables set to 'true' during the observation period. As noted earlier, this corresponds to the number of CSP variables that would be bound in the direct CSP encoding.

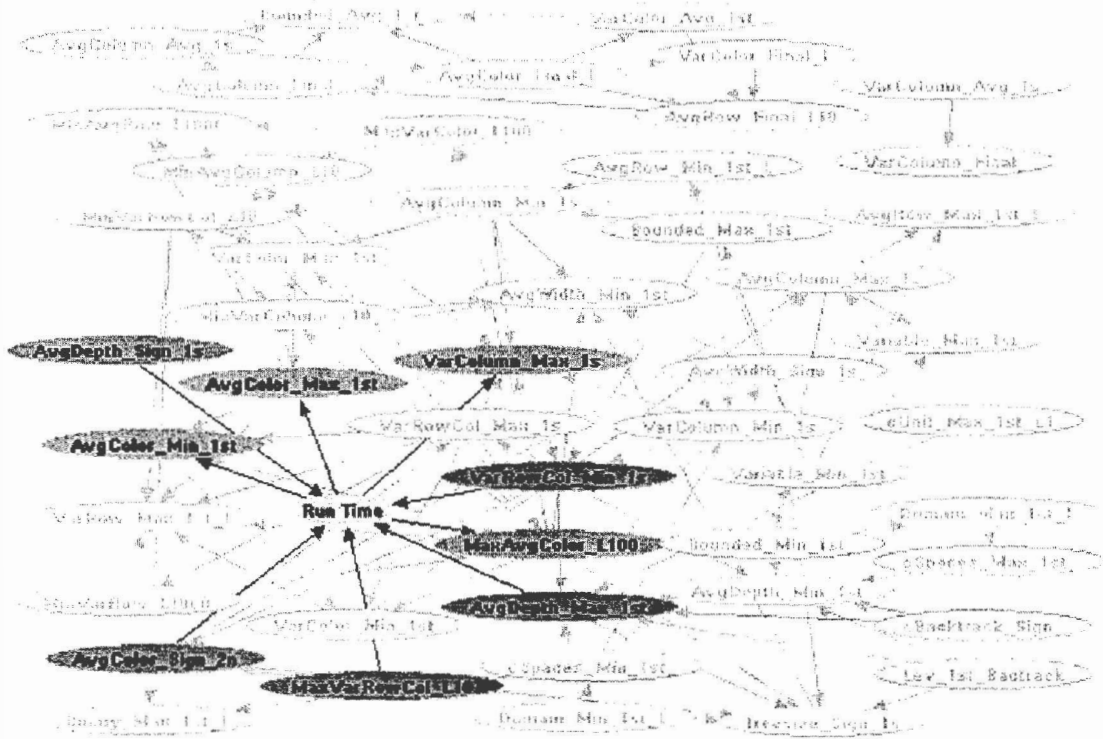

Figure 3: The learned Bayesian network for a sample CSP-QWH-Single problem. Key dependencies and variables are highlighted.

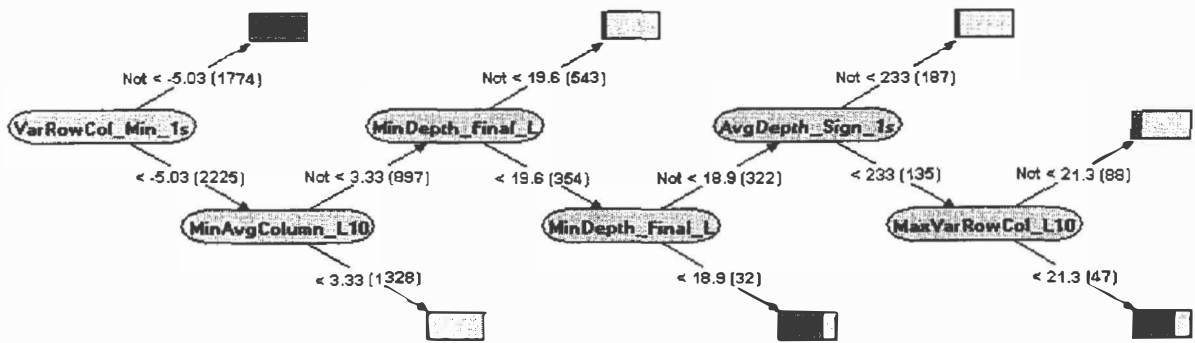

Figure 4: The decision tree inferred for run time from data gathered in a CSP-QWH-Single experiment. The probability of a short run is captured by the light component of the bargraphs displayed at the leaves.

- The number of models ruled out.

- The number of unit propagations performed.

- The number of variables eliminated by Satz's lookahead component: that is, the effectiveness of lookahead.

- The quantity $\lambda$ described in Sec. 3.1 above, a measure of the constrainedness of the binary clause subproblem.

### 4.4 Satz-QWH-Multi Problem

For the experiment with the Satz-QWH-Multi problem, we executed single runs of QWH instances with the same parameters as the instance studied in the Satz-QWH-Single Problem (bqwh-34-410) for the training and test sets. Run time and observational variables were normalized in the same manner as for the CSP-QWH-Multi problem. The classification accuracy of the learned model was found to be 0.715. The classification accuracy of the marginal model was found to be 0.526. The average log score for the model was -0.557 and the average log score for the marginal model was -0.692.

### 4.5 Toward Larger Studies

For broad application in guiding computational problem solving, it is important to develop an understanding of how results for sample instances, such as the problems described in Sections 4.1 through 4.4, generalize to new instances within and across distinct classes of problems. We have been working to build insights about generalizability by exploring the statistics of the performance of classifiers on sets of problem instances. The work on studies with larger numbers of data sets has been limited by the amount of time required to generate data sets for the hard problems being studied. With our computing platforms, several days of computational effort were typically required to produce each data set.

As an example of our work on generalization, we review the statistics of model quality and classification accuracy, and the regularity of discriminatory features for additional data sets of instances in the CSP-QWH-Single problem class.

We defined ten additional nonbalanced QWH problem instances, parameterized in the same manner as the CSP problem described in Section 4.1 (order 34 with 380 unassigned variables). We employed the same data generation and analysis procedures as before, building and testing ten separate models. Generating data for these analyses using the ILOG libary executed on an Intel Pentium III (running at 600 Mhz) required approximately twenty-four hours per 1000 runs. Thus, each CSP dataset required approximately five days of computation.

In summary, we found significant boosts in classification accuracy for all of the instances. For the ten datasets, the mean classification accuracy for the learned models was 0.812 with a standard deviation of 0.101. The average log score for the models was -0.388 with a standard deviation of 0.167. The predictive power of the learned models stands in contrast to the classification accuracy of using background statistics; the mean classification accuracy of the marginal models was 0.497 with a standard deviation of 0.025. The average log score for the marginal models was -0.693 with a standard deviation of 0.001. Thus, we observed relatively consistent predictive power of the methods across the new instances.

We observed variation in the tree structure and discriminatory features across the ten learned models. Nevertheless, several features appeared as valuable discriminators in multiple models, including statistics based on measures of VarRowColumn, AvgColumn, AvgDepth, and MinDepth. Some of the evidential features recurred for different problems, showing significant predictive value across models with greater frequency than others. For example, measures of the maximum variation in the number of uncolored cells in the QWH instance across rows and columns (MaxVarRowColumn) appeared as being an important discriminator in many of the models.

## 5 Generalizing Observation Policies

For the experiments described in Sections 3 and 4, we employed a policy of gathering evidence over an observation horizon of the initial 1000 choice points. This observational policy can be generalized in several ways. For example, in addition to harvesting evidence within the observation horizon, we can consider the amount of *time expended so far* during a run as an explicit observation. Also, evidence gathering can be generalized to consider the status of variables and statistics of variables at progressively later times during a run.

Beyond experimenting with different observational policies, we believe that there is potential for harnessing value-of-information analyses to optimize the gathering of information. For example, there is opportunity for employing offine analysis and optimization to generate tractable real-time observation policies that dictate which evidence to evaluate at different times during a run, conditioned on evidence that has already been observed during that run.

### 5.1 Time Expended as Evidence

In the process of exploring alternate observation policies, we investigated the value of extending the bounded-horizon policy described in Section 3, with a consideration of the status of time expended so far during a run. To probe potential boosts with inclusion of time expended, we divided several of the data sets explored in Section 4.5 into subsets based on whether runs with the data set had exceeded specific run-time boundaries. Then, we built distinct run-time–specific models and tested the predictive power of these models on test sets containing instances of appropriate minimal length. Such time-specific models could be used in practice as a cascade of models, depending on the amount of time that had already been expended on a run.

We typically found boosts in the predictive power of models built with such temporal decompositions. As we had expected, the boosts are greatest for models conditioned on the largest amounts of expended time. As an example, let us consider one of the data sets generated for the study in Section 4.5. The model that had been built previously with all of the data had a classification accuracy of 0.793. The median time for the runs represented in the set was nearly 18,000 choice points. We created three separate subsets of the complete set of runs: the set of runs that exceeded 5,000 choice points, the set that exceeded 8,000 choice points, and the set that had exceeded 11,000 choice points. We created distinct predictive models for each training set and tested these models with cases drawn from test sets containing runs of appropriate minimal length. The classification accuracies of the models for the low, medium, and high time expenditure were 0.779, 0.799, and 0.850 respectively. We shall be continuing to study the use of time allocated as a predictive variable.

## 6 Application: Dynamic Restart Policies

A predictive model can be used in several ways to control a solver. For example, the variable selection heuristic used to decompose the problem instance can be designed to minimize the expected solution time of the subproblems. Another application centers on building distinct models to predict the run time associated with different global strategies. As an example, we can learn to predict the relative performance of ordinary chronological backtrack search and dependency-directed backtracking with clause learning [16]. Such a predictive model could be used to decide whether the overhead of clause learning would be worthwhile for a particular instance.

Problem and instance-specific predictions of run time can also be used to drive dynamic cutoff decisions on when to suspend a current case and restart with a new random seed or new problem instance, depending on the class of problem. For example, consider a greedy analysis, where we deliberate about the value of ceasing a run that is in progress and performing a restart on that instance or another instance, given predictions about run time. The predictive models described in this paper can provide the expected time remaining until completion of a current run. Initiating a new run will have an expected run time provided by the statistics of the marginal model. From the perspective of a single-step analysis, when the expected time remaining for the current instance is greater than the expected time of the next instance, as defined by the background marginal model, it is better to cease activity and perform a restart. More generally, we can construct richer multistep analyses that provide the fastest solutions to a particular instance or the highest rate of completed solutions with computational effort.

We can also use the predictive models to perform comparative analyses with previous policies. Luby et al. [21] have shown that the optimal restart policy, assuming full knowledge of the distribution, is one with a fixed cutoff. They also provide a universal strategy (using gradually increasing cutoffs) for minimizing the expected cost of randomized procedures, assuming no prior knowledge of the probability distribution. They show that the universal strategy is within a log factor of optimal. These results essential settle the distribution-free case.

Consider now the following dynamic policy: Observe a run for $O$ steps. If a solution is not found, then predict whether the run will complete within a total of $L$ steps. If the prediction is negative, then immediately restart; otherwise continue to run for up to a total of $L$ steps before restarting if no solution is found.

An upper bound on the expected run of this policy can be calculated in terms of the model accuracy $A$ and the probability $P_i$ of a single run successfully ending in $i$ or fewer steps. For simplicity of exposition we assume that the model's accuracy in predicting long or short runs is identical. The expected number of runs until a solution is found is $E(N) = 1/(A(P_L - P_O) + P_O)$. An upper bound on the expected number of steps in a single run can be calculated by assuming that runs that end within $O$ steps take exactly $O$ steps, and that runs that end in $O+1$ to $L$ steps take exactly $L$ steps. The probability that the policy continues a run past $O$ steps (i.e., the prediction was positive) is $AP_L + (1-A)(1-P_L)$. An upper bound on the expected length of a single run is $E_{ub}(R) = O + (L-O)(AP_L + (1-A)(1-P_L))$. Thus, an upper bound on the expected time to solve a problem using the policy is $E(N)E_{ub}(R)$.

It is important to note that the expected time depends on both the accuracy of the model and the prediction point $L$; in general, one would want to vary $L$ in order to optimize the solution time. Furthermore, in general, it would be better to design more sophisticated dynamic policies that made use of all information gathered over a run, rather than just during the first $O$ steps. But even a non-optimized policy based directly on the models discussed in this paper can outperform the optimal fixed policy. For example, in the CSP-QWH-single problem case, the optimal fixed policy has an expected solution time of 38,000 steps, while the dynamic policy has an expected solution time of only 27,000 steps. Optimizing the choice of $L$ should provide about an order of magnitude further improvement.

While it may not be surprising that a dynamic policy can outperform the optimal fixed policy, it is interesting to note that this can occur when the observation time $O$ is *greater* than the fixed cutoff. That is, for proper values of $L$ and $A$, it may be worthwhile to observe each run for 1000 steps even if the optimal fixed strategy is to cutoff after 500 steps. These and other issues concerning applications of prediction models to restart policies are examined in detail in a forthcoming paper.

## 7 Related Work

Learning methods have been employed in previous research in a attempt to enhance the performance optimize reasoning systems. In work on "speed-up learning," investigators have attempted to increase planning efficiency by learning goal-specific preferences for plan operators [22, 19]. Khardon and Roth explored the offline reformulation of representations based on experiences with problem solving in an environment to enhance run-time efficiency [18]. Our work on using probabilistic models to learn about algorithmic performance and to guide problem solving is most closely related to research on flexible computation and decision-theoretic control. Related work in this arena focused on the use of predictive models to control computation, Breese and Horvitz [3] collected data about the progress of search for graph cliquing and of cutset analysis for use in minimizing the time of probabilistic inference with Bayesian networks. The work was motivated by the challenge of identifying the ideal time for preprocessing graphical models for faster inference before initiating inference, trading off reformulation time for inference time. Trajectories of progress as a function of parameters of Bayesian network problem instances were learned for use in dynamic decisions about the partition of resources between reformulation and inference. In other work, Horvitz and Klein [14] constructed Bayesian models considering the time expended so far in theorem proving. They monitored the progress of search in a propositional theorem prover and used measures of progress in updating the probability of truth or falsity of assertions. A Bayesian model was harnessed to update belief about different outcomes as a function of the amount of time that problem solving continued without halting. Stepping back to view the larger body of work on the decision-theoretic control of computation, measures of *expected value of computation* [15, 8, 25], employed to guide problem solving, rely on forecasts of the refinements of partial results with future computation. More generally, representations of problem-solving progress have been central in research on flexible or anytime methods—procedures that exhibit a relatively smooth surface of performance with the allocation of computational resources.

## 8 Future Work and Directions

This work represents a vector in a space of ongoing research. We are pursuing several lines of research with the goals of enhancing the power and generalizing the applicability of the predictive methods. We are exploring the modeling of run time at a finer grain through the use of continuous variables and prototypical named distributions. We are also exploring the value of decomposing the learning problem into models that predict the average execution times seen with multiple runs and models that predict how well a particular instance will do relative to the overall hardness of the problem. In other extensions, we are exploring the feasibility of inferring the likelihood that an instance is solvable versus unsolvable and building models that forecast the overall expected run time to completion by conditioning on each situation. We are also interested in pursuing more general, dynamic observational policies and in harnessing the value of information to identify a set of conditional decisions about the pattern and timing of monitoring. Finally, we are continuing to investigate the formulation and testing of ideal policies for harnessing the predictive models to optimize restart policies.

## 9 Summary

We presented a methodology for characterizing the run time of problem instances for randomized backtrack-style search algorithms that have been developed to solve a hard class of structured constraint-satisfaction problems. The methods are motivated by recent successes with using fixed restart policies to address the high variance in running time typically exhibited by backtracking search algorithms. We described two distinct formulations of problem-solving goals and built probabilistic dependency models for instances in these problem classes. Finally, we discussed opportunities for leveraging the predictive power of the models to optimize the performance of randomized search procedures by monitoring execution and dynamically setting cutoffs.

## Acknowledgments

We thank Dimitris Achlioptas for his insightful contributions and feedback.